%% file: main.tex
\pdfoutput=1
\documentclass[11pt]{article}

\usepackage{EMNLP2022}

\usepackage{times}
\usepackage{latexsym}
\usepackage[T1]{fontenc}
\usepackage[utf8]{inputenc}
\usepackage{microtype}
\usepackage{inconsolata}

\usepackage{graphicx}
\usepackage{xcolor}
\usepackage{comment}
\usepackage{booktabs}
\usepackage{amsmath}
\usepackage{amssymb}
\usepackage{amsthm}
\usepackage{bbm} % indicator function
\usepackage{subcaption}
\usepackage{multirow}
\usepackage{comment}
\usepackage{makecell} %multiline cell
\usepackage{url}
\usepackage{hyperref}
\usepackage{tabularx}
\usepackage[normalem]{ulem}
\useunder{\uline}{\ul}{}

\usepackage[export]{adjustbox}
\usepackage{paralist}

\title{
Zero-Shot Learners for Natural Language Understanding via a Unified Multiple Choice Perspective
}

\author{
    Ping Yang$^\textnormal{1}$\footnotemark[1] \qquad
    Junjie Wang$^\textnormal{2}$\footnotemark[1] \qquad
    Ruyi Gan$^\textnormal{1}$ \qquad
    Xinyu Zhu$^\textnormal{3}$ \qquad
    Lin Zhang$^\textnormal{1}$ \\
    \textbf{
    Ziwei Wu$^\textnormal{1}$ \qquad
    Xinyu Gao$^\textnormal{1}$ \qquad
    Jiaxing Zhang$^\textnormal{1}$\ \qquad
    Tetsuya Sakai$^\textnormal{2}$\footnotemark[2]
    }
    \\
    $^\textnormal{1}$International Digital Economy Academy \quad
    $^\textnormal{2}$Waseda University \quad
    $^\textnormal{3}$Tsinghua University \\
    {\tt\small \{yangping, ganruyi, zhanglin, wuziwei, gaoxinyu, zhangjiaxing\}@idea.edu.cn } \\
    {\tt\small wjj1020181822@toki.waseda.jp} \quad {\tt\small tetsuyasakai@acm.org} \quad {\tt\small zhuxy21@mails.tsinghua.edu.cn }
}

\begin{document}
\maketitle

{
  \renewcommand{\thefootnote}%
    {\fnsymbol{footnote}}
  \footnotetext[1]{Equal contribution.}
  \footnotetext[2]{Corresponding Author.}
}

\begin{abstract}
We propose a new paradigm for zero-shot learners that is format agnostic, i.e., it is compatible with any format and applicable to a list of language tasks, such as text classification, commonsense reasoning, coreference resolution, and sentiment analysis.
Zero-shot learning aims to train a model on a given task such that it can address new learning tasks without any additional training.
Our approach converts zero-shot learning into multiple-choice tasks, avoiding problems in commonly used large-scale generative models such as FLAN.
It not only adds generalization ability to models but also significantly reduces the number of parameters.
Our method shares the merits of efficient training and deployment.
Our approach shows state-of-the-art performance on several benchmarks and produces satisfactory results on tasks such as natural language inference and text classification.
Our model achieves this success with only 235M parameters, which is substantially smaller than state-of-the-art models with billions of parameters.
The code and pre-trained models are available at \url{https://github.com/IDEA-CCNL/Fengshenbang-LM/tree/main/fengshen/examples/unimc}.
\end{abstract}

\section{Introduction}
\input{sections/01introduction}

\section{Related Work}
\input{sections/02related}

\section{Approaches}
\input{sections/03framework.tex}

\section{Experiments}
\input{sections/04experiment.tex}

\section{Conclusions}
\input{sections/05conclusion}

\section*{Limitations}
\input{sections/06limitation}

\section*{Ethical Considerations}
\input{sections/07ethic}

\section*{Acknowledgements}
\input{sections/ack}

\small{
\bibliography{anthology, custom}
\bibliographystyle{acl_natbib}
}

\input{sections/appendix.tex}

\end{document}

%% file: sections/01introduction.tex
\begin{figure}[!htb]
    \centering
    \includegraphics[width=0.48\textwidth]{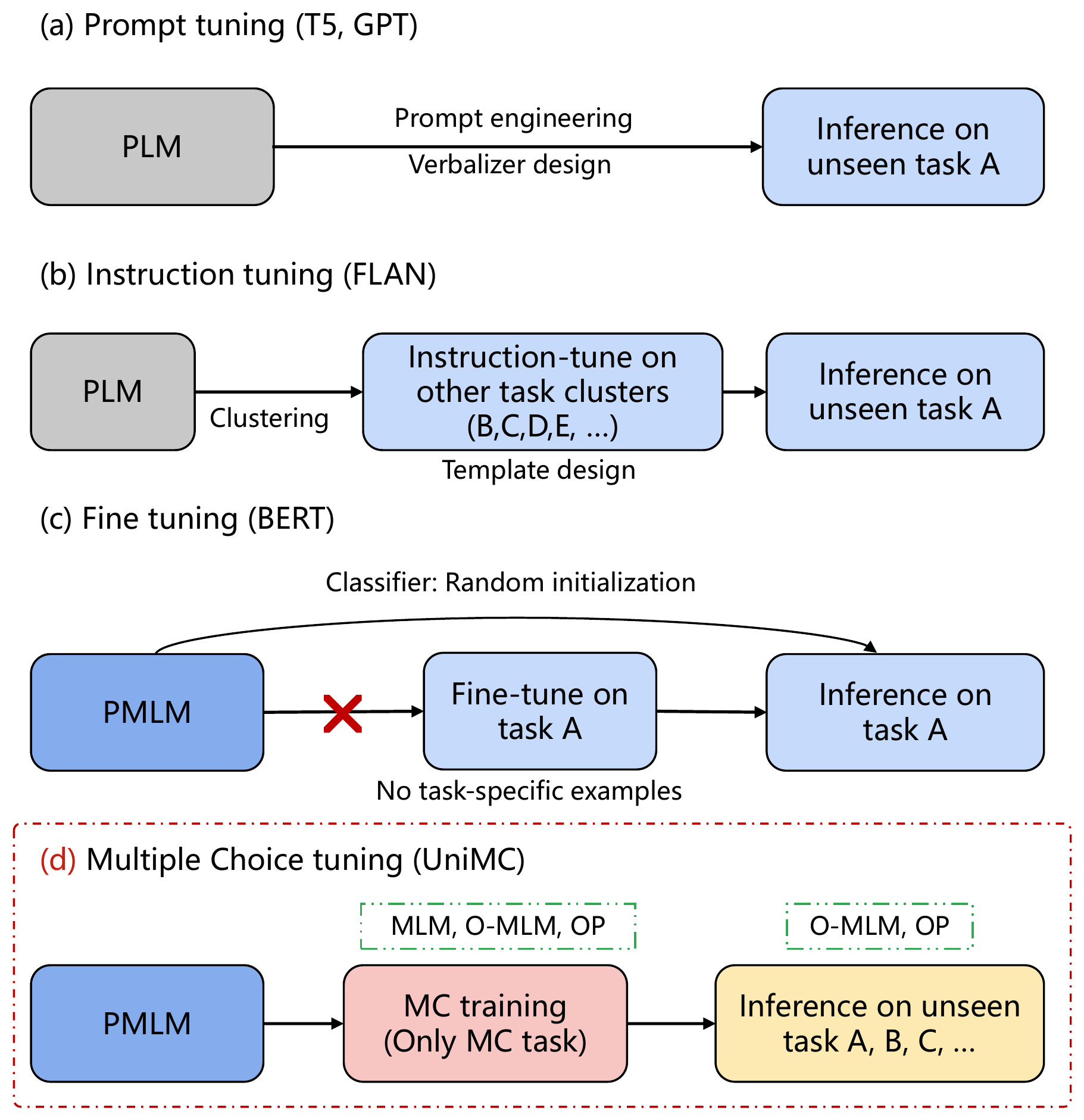}
    \caption{
    Typical zero-shot learning methods and our proposed UniMC. ``PLM'' indicates pre-trained language model. ``PMLM'' implies pre-trained masked language model.
    }
    \label{fig:common_learning_vs_multitask}
\end{figure} 

\begin{figure}[!htb]
    \centering
    \includegraphics[width=0.45 \textwidth]{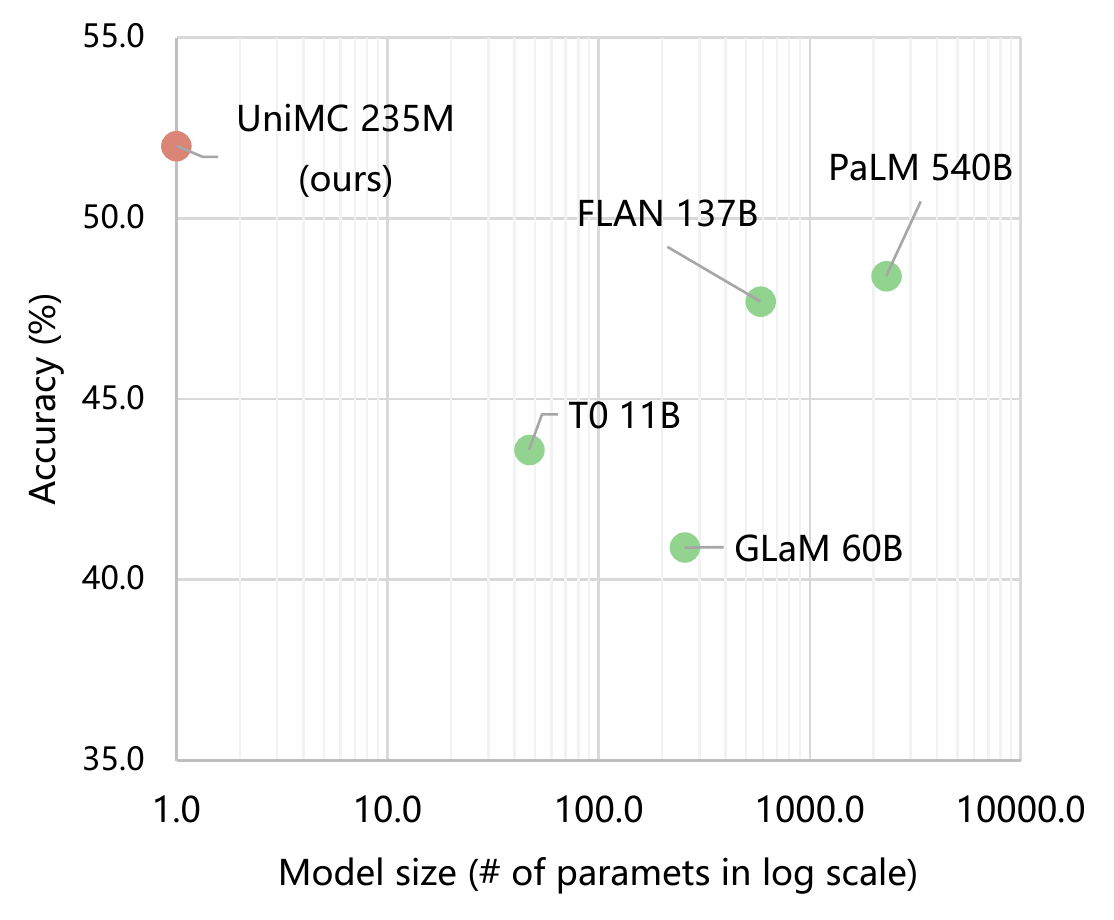}
    \caption{
    Zero-shot performance comparison in ALNI R1. Our proposed UniMC has the best performance w.r.t the accuracy and the model size, simultaneously.
    }
    \label{fig:nli_and_parameter}
\end{figure}

% =================================================

Remarkable advances in large-scale language models have brought substantial improvements in a wide variety of tasks such as text classification, natural language inference and commonsense reasoning~\cite{DBLP:conf/nips/BrownMRSKDNSSAA20/gpt3, DBLP:journals/corr/abs-2204-02311/PaLM}.
This progress brings opportunities to Zero-Shot Learning (ZSL)~\cite{DBLP:journals/corr/abs-2110-08207/T0,DBLP:journals/corr/abs-2204-02311/PaLM}, which aims to predict labels on datasets from novel domains. 
Most solutions can be framed in the prompt tuning framework that activate specific parameters in PLM~\cite{DBLP:journals/corr/abs-2201-06910/zeroprompt,DBLP:journals/corr/abs-2103-10385/p-tuning} to adapt to zero-shot tasks. 
A powerful variant of prompt tuning is called instruction tuning~\cite{Wei2021FinetunedLM/FLAN}, which shares knowledge from different domains.
We summarize the mainstream large-scale frameworks in Figure~\ref{fig:common_learning_vs_multitask}.

Despite their success, these frameworks suffer from their inherent problems, and thus limit their potential in zero-shot learners.
Firstly, prompt-related models have an extremely large number of parameters, e.g., GPT-3 has 175B, FLAN has 137B and PaLM~\cite{DBLP:journals/corr/abs-2204-02311/PaLM} has 540B. 
One immediate problem is that these models are often hard to be trained, making the deployment and consumption difficult.
Secondly, manual processing is required when addressing zero-shot problems.
For instance, T0 builds $2,073$ prompts to handle more than $170$ tasks~\cite{DBLP:journals/corr/abs-2110-08207/T0}. 
Lastly, existing models employ a single direction paradigm, either auto-regressive models or sequence-to-sequence, resulting in inadequate usage of information from both directions. 
As an example, PMLM tries to implement a zero-shot learner, 
which is shown in Figure~\ref{fig:common_learning_vs_multitask} (c).
Note that recent work~\cite{DBLP:conf/acl/LiuHCG19/nlu} state that PMLM is more suitable than PLM for Natural Language Understanding (NLU) tasks.
However, it has to be fine-tuned on the task-specific samples to initialize the classifier instead of randomly initializing the classifier.
Therefore, the ability of PMLM is limited when dealing with zero-shot scenarios.

To address the aforementioned problems, we introduce a light-weight framework, called  \textbf{Uni}fied \textbf{M}ultiple \textbf{C}hoice model (UniMC), proposing a novel MC tuning. 
The proposed MC tuning has the following advantages: 
i) parameter updating only happens in the MC training phase, and ii) facilitating the deployment. 
To reduce the manual processing, we only formulate one candidate option prompt format and one question prompt format. 
Note that we also consider the case without any question prompt format.
Under this setting, we can treat labels as options rather than building verbalizer maps and providing its text information to the models as before. 
We therefore can learn the information from labels directly. 
To this end, we convert the problematic classifiers to options.
One immediate question is how to choose an option efficiently and unambiguously.
Therefore, as shown in Section~\ref{sec:mc_tuning}, we develop an option-mask tokens {\tt [O-MASK]} to predict ``yes'' or ``no'' before each option. 
A two-step process is introduced to output the desired options.
First, similar to Masked Language Modeling (MLM)~\cite{DBLP:conf/naacl/DevlinCLT19/bert}, we conduct Option MLM (O-MLM) to recover the ``yes'' or ``no'' for each option. 
Next, we propose an Option Prediction (OP) method to compute proper options. 

With extensive experiments on multiple challenging benchmarks, we demonstrate that our approach's performance outperforms state-of-the-art baselines, while reducing the model size with two orders, as shown in Figure~\ref{fig:nli_and_parameter}. 
This success suggests the potential of leveraging UniMC in large datasets.
Our contributions are as follows.
\begin{itemize}
    \item[$\bullet$]
    We propose a new zero-shot paradigm by converting this problem into multiple choice tasks.
    \item[$\bullet$]
    We develop an effective and efficient method to implement a MC-based zero-shot learner. Our proposed method has up to $48\%$ improvement on Dbpedia over SOTA baselines that have a few hundred times larger than our model.
\end{itemize}

%% file: sections/02related.tex
\subsection{Unified NLP Task Formats}

NLP tasks often have diverse formats due to the fast emergence of datasets, such as machine reading comprehension and text classification tasks. 
Recent research shows the necessity of unifying formats to fix the gap across various tasks~\cite{DBLP:journals/corr/abs-2110-08207/T0, Wei2021FinetunedLM/FLAN,DBLP:journals/corr/abs-2109-03564/NSP-BERT}. 
By developing a natural language prompted form, T0 builds an application to map original NLP datasets into target templates with custom prompts~\cite{DBLP:journals/corr/abs-2110-08207/T0}. 
FLAN groups multiple datasets into $12$ task clusters, and then designs $10$ unique instruction templates to unify formats~\cite{Wei2021FinetunedLM/FLAN}. 
Despite effective, this focuses on generative styles and thus cannot be adapted to vast label-based models that select.
This motivates us to unify label-based tasks, where we develop unified Multiple Choice (MC) formats for this purpose.

\subsection{Label Information}

The label semantic is an important information source, such as in few-shot tasks~\cite{DBLP:conf/acl/HouCLZLLL20/cdt,DBLP:conf/acl/MuellerKRMMZR22/lsap,DBLP:conf/acl/LuoLLZ21/lasaml}.
The L-TapNet framework~\cite{DBLP:conf/acl/HouCLZLLL20/cdt} integrates the label information with manually designed prompts for inputs to solve few-shot slot tagging tasks.
In addition, LSAP~\cite{DBLP:conf/acl/MuellerKRMMZR22/lsap} obtains powerful few-shot performance by introducing label semantics into the pre-training and fine-tuning phases of the PLMs.
Together, these successful employments of labels in low-resource settings inspire us to bring label semantics to our unified MC inputs to handle the zero-shot scenario.

\subsection{Zero-Shot Learning}
Large-scale Pre-trained Language Models (PLMs) with billions of parameters such as GPT-3~\cite{DBLP:conf/nips/BrownMRSKDNSSAA20/gpt3} have shown impressive performance across various few-shot tasks. 
However, they have limited competence when dealing with zero-shot tasks, which have broader applications in practice. 
Recent efforts try to  mitigate this issue from different perspectives. 
FLAN~\cite{Wei2021FinetunedLM/FLAN} designs specific instruction templates for each task and utilizes over $60$ labeled datasets to ``fine-tune'' a 137B language model. 
T0~\cite{DBLP:journals/corr/abs-2110-08207/T0} unifies all tasks into a source-target format by collecting a large variety of prompt templates, specifically $2,073$ manually constructed prompts, and trains the model with multi-task learning. 
Along this line, ZeroPrompt~\cite{DBLP:journals/corr/abs-2201-06910/zeroprompt} applies over $1,000$ supervised datasets and proposes the genetic prompt search method to find prompts for new tasks. 
However, these methods cost significant laborious efforts, such as  prompt engineering and template designing. 
Moreover, the pre-training and tuning phases of large-scale PLMs take enormous amounts of computational resources, therefore, new tasks may suffer great difficulty in deploying. 
As a comparison, our proposed UniMC is light-weighted, i.e., has 235M parameters and a few manual input text transformations, making it suitable for more general scenarios.

%% file: sections/03framework.tex
\begin{figure}[t]
    \centering
    \includegraphics[width=0.5\textwidth]{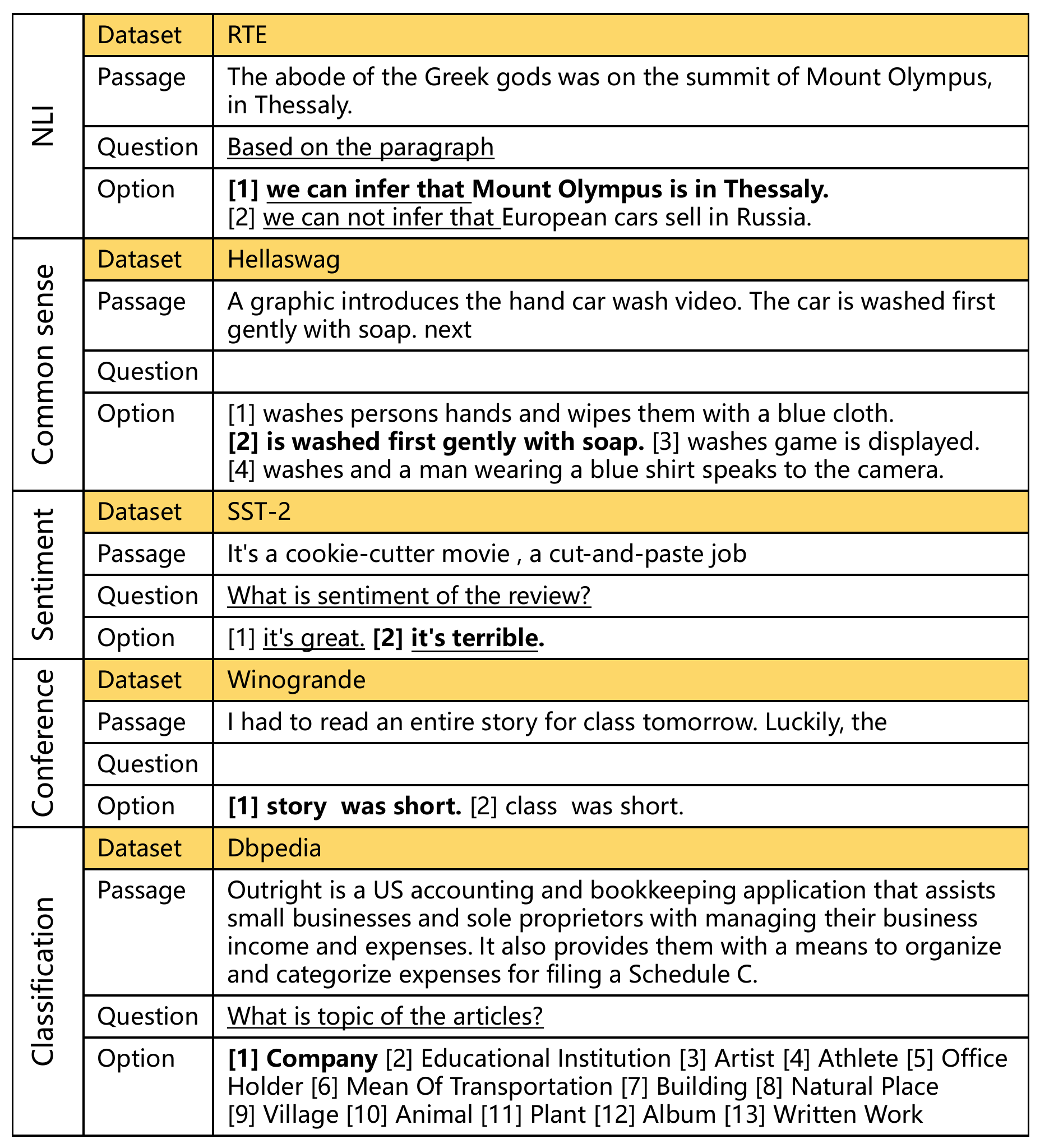}
    \caption{Unified input text examples with sampling from datasets in zero-shot phase. The prompt text is \underline{underlined} and the correct options are in \textbf{bold}.}
    \label{fig:examples_zero_shot}
\end{figure} 

In this section, we outline the proposed framework, i.e., UniMC, and provide the training and inference approaches in detail.

\subsection{The UniMC framework}
\label{sec:unimc_framework}

\subsubsection{Unified Input}
\label{sec:unified_input}

A unified input format will facilitate the generalization of models, promoting the sharing of knowledge across different tasks.
To achieve this, we frame all tasks' objectives together as a multiple-choice (MC) problem, as shown in Figure~\ref{fig:examples_zero_shot}.
A MC problem often consists of three components, including  options, question, and passage. 
We now discuss the details of getting these bodies. 
We can often get the passage component effortlessly because it often exists in the original data.
As to the question part, we can either use the raw question directly or provide a corresponding question when it is missing.
The transformation of options depends on whether or not we can get a straightforward expression of classes.
On the one hand, we can convert all classification tasks into options directly as it has specific information for choices.
On the other hand, we have to construct an option prompt to generate particular choices.
Details of this transformation can be found in Appendix~\ref{append:dataset_details}. 
In effect, these allow us to abandon label indices as in classification tasks, which include much less information than our used options. 

% In detail，passage通常直接来自于原来的数据中的passage（因为数据集中都会有passage所以就可以直接用），和question一同构成了context information。
% 然而对于question，假如问题是在数据集中存在的，那么我们和passage一样，直接用原来的文字就好了。
% 假如有一些特定的问题是缺失的话，我们就只加入一个简单的question prompt帮助人类/模型对本文句的理解。
% 相似的，对于option来说，比如像是classification任务，有明显的文字作为类别（只不过是以label的形式存在的），我们就可以直接使用，作为option。
% 假如，这些选择是某种隐式的表达，比如在sentiment任务中，是需要判断这是积极的还是消极的，数据集中并没有使用一个准确的“自然语言”来进行描述，取而代之的是一个“数字”，那么我们就会构造一个option prompt给模型进行选择。

\begin{figure*}[!htb]
    \centering
    \includegraphics[width=\textwidth]{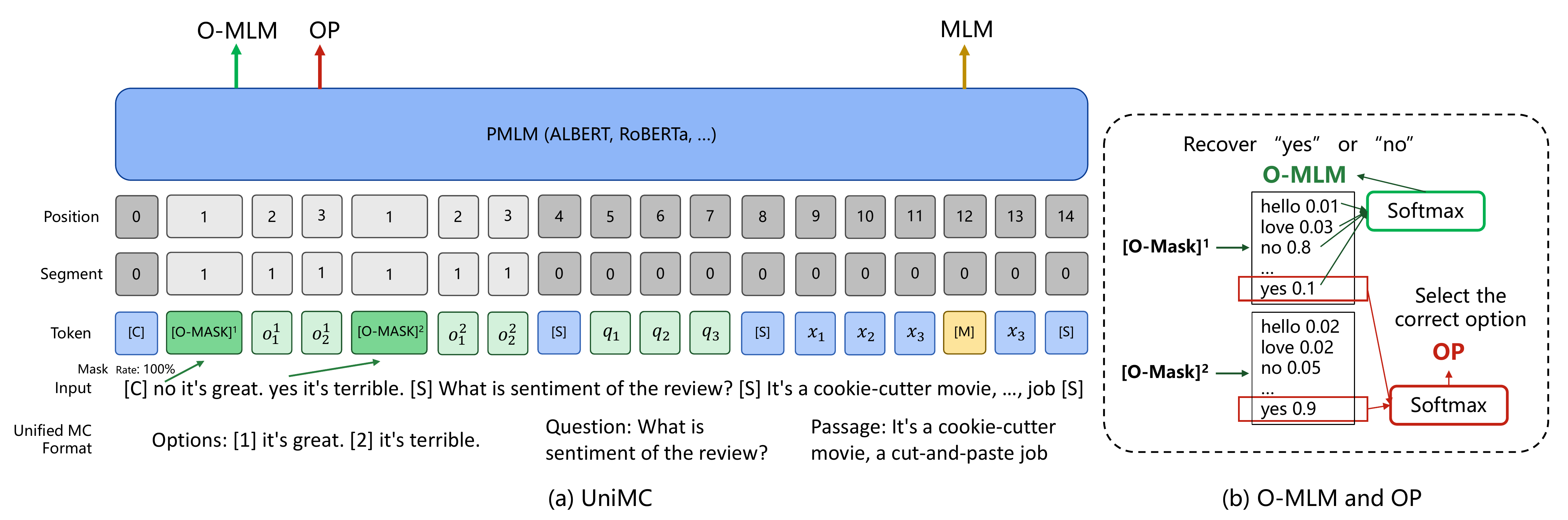}
    \caption{UniMC framework with O-MLM and OP in MC training phase. ``{\tt [O-MASK]$^1$}'' in (b) indicates the option mask token of option 1. Similarly, ``{\tt [O-MASK]$^2$}'' is related to option 2. {\tt [C]}, {\tt [S]} and {\tt [M]} are the abbreviation of {\tt [CLS]}, {\tt [SEP]} and {\tt [MASK]}. The example of input text is from the dataset SST-2~\cite{DBLP:conf/emnlp/SocherPWCMNP13/sst-2}.}
    \label{fig:overview_unimc}
\end{figure*}

\subsubsection{Network}
\label{sec:backbone_network}

In our framework, we employ BERT-like PMLMs as the backbone, such as ALBERT~\cite{DBLP:conf/iclr/LanCGGSS20/albert} and RoBERTa~\cite{DBLP:journals/corr/abs-1907-11692/roberta}, to integrate the bidirectional modeled input $x_{inp}$. 
In additional, the discussion of backbone models is in Appendix~\ref{append:experiment_details}. 
Instead of using the original embedding methods directly, we develop a new solution for the segment id, position id, and attention mask matrix to fit multiple choice tasks, simultaneously. 

{\noindent \bf Tokenization:}
In this framework, the key to achieve the ability of addressing MC tasks is to set up a proper option. 
We thus introduce option-mask tokens ({\tt [O-MASK]}), aiming to replace ``yes'' or ``no'' in the input text for a better representation ability. % 应该理解为：表示能力 representation
Here, {\tt [O-MASK]} inherits the ability of {\tt [MASK]}, and thus remains to use token predictions to determine which option is correct.
Consider, as an example, an input set, denoted as $(o, q, x)$, includes the following: i) one passage $x=x_1 \dots x_{\left| x \right|}$,  ii) $N_Q$ questions $q=q_1 \dots q_{\left| q \right|}$, and iii) $N_O$ candidate options $o = o_1 \dots o_{\left| o \right|} $, whose  input token $x_{inp}$ is formulated as follows:
\begin{equation}
\small
\begin{aligned}
x_{inp} = &\text{[CLS]} \{ \text{[O-MASK]}^i \,\, o^i \}_{i=1}^{N_O} \text{[SEP]} \\
            &\{ q \,\, \text{[SEP]}\}^{N_Q} \,\, x \,\, \text{[SEP]},
\end{aligned}
\end{equation}
Here, $N_Q \in \{ 0, 1 \}$, $N_O \in \mathbb{N+}$ and $N_O \geq 2$. 

\begin{figure}[!tp]
    \centering
    \includegraphics[width=0.46\textwidth]{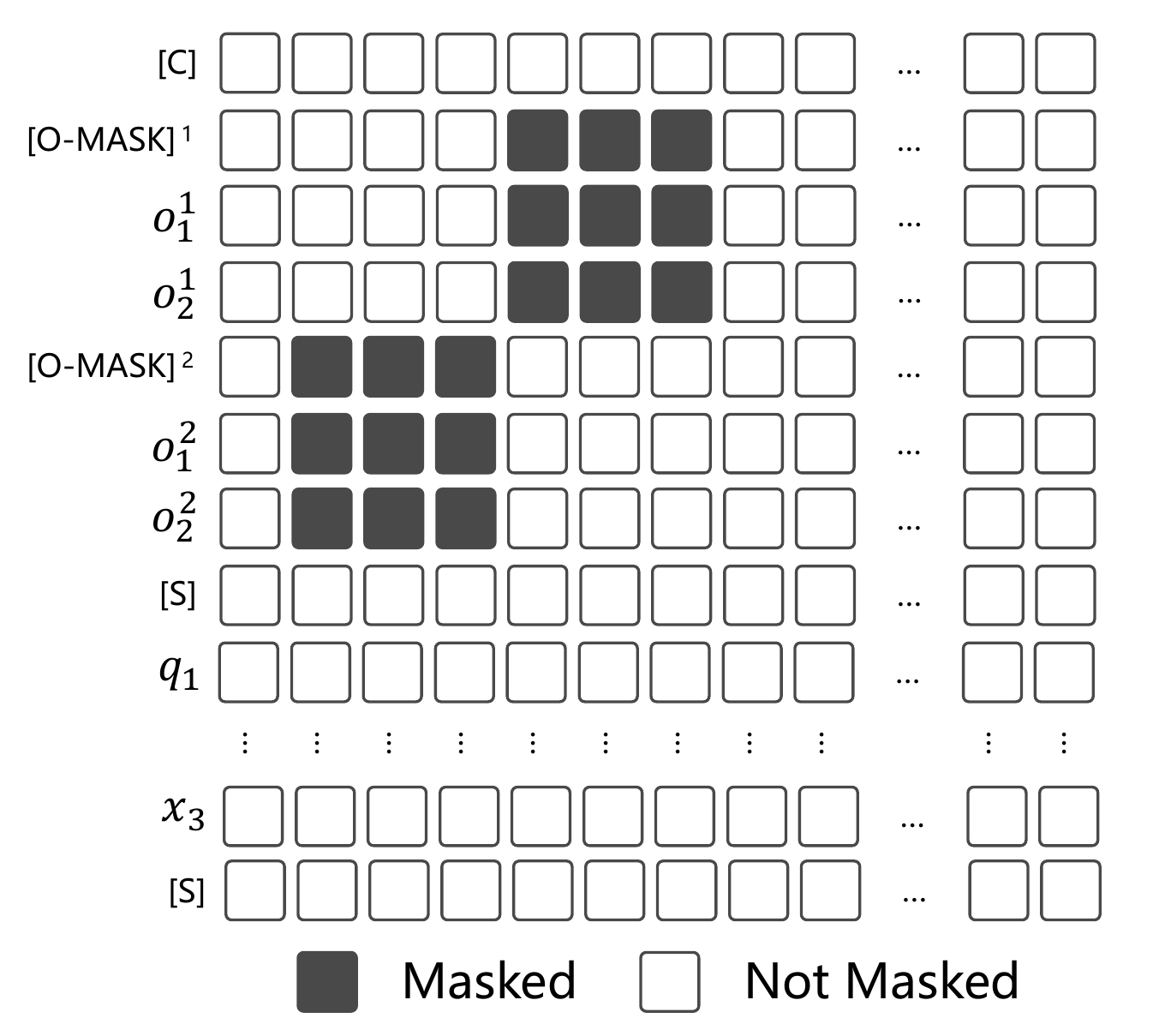}
    \caption{Self-Attention Mask Matrix. Given input ${\tt [C]}, {\tt [O-MASK]}, o_1^1, o_2^1, \dots, x_3, {\tt [S]}$, the tokens of options can not attend to each other.}
    \label{fig:self_attention_mask}
\end{figure} 

{\noindent \bf Id embeddings and attention mask matrix: }
Note that a unified input text has multiple options, leading to undesired mutual influence between options and resulting in a misunderstanding of answers.
We now address this issue from the following three perspectives, including segment id, position id, and attention mask matrix.
Firstly, we assign segment id to distinguish option and context (questions, passages) information, as shown in Fig.~\ref{fig:overview_unimc} (a). 
Secondly, we update the position id to tell apart the intra information in the option. 
This is because that PMLMs cannot get position information from tokens. 
We aim to allow PMLMs will treat tokens' position information based on their position embeddings. 
Lastly, we will control the flow between options, such as $M_{mask}$ in self-attention, as shown in Fig.~\ref{fig:self_attention_mask}.
In particular, black squares are used to mask a part of the input attention matrix, ensuring the disentanglement between different options.
We place a $-inf$ number on the masked slots, which is the same as BERT to mask tokens.

Furthermore, we can have the encoded hidden vector, denoted as  $T=[T_1 \dots T_n]$, using multiple Transformer-based layers as following,
\begin{equation}
\small
T=\operatorname{encoder}(x_{inp}, pos, seg, M_{mask}).
\end{equation}

\subsection{MC tuning}
\label{sec:mc_tuning}

Recall the backbones are often pre-trained models, resulting in excellent skill in capturing the commonsense knowledge.
Intuitively, we can employ these as base modules by taking advantage of their high volume knowledge.
More specifically, we use the outputs of pre-trained models as the initial states for the following MC tasks, leading to a two-stage tuning paradigm.
In the MC training phase, we train the models with MC tasks and gain a great initialization for selecting a correct option. 
In the zero-shot phase, we apply the unified MC models to unseen zero-shot tasks.

\subsubsection{MC training phase} 

We now introduce the proposed option masked language modeling (O-MLM) and option prediction (OP) methods in detail. 

Masked Language Modeling (MLM) is a pre-training task in BERT~\cite{DBLP:conf/naacl/DevlinCLT19/bert} for self-supervised learning,
\begin{equation}
\small
\mathcal{L}_{\mathrm{MLM}}=-\sum_{\hat{T} \in m(T)} \log p\left(\hat{T} \mid T_{\backslash m(T)}\right),
\end{equation}
where $\hat{T}$ is the random perturbed token from $T$;
$m(T)$ and $T_{\backslash m(T)}$ are the masked tokens from $T$ and the reset tokens, respectively.
In practice, we randomly replace tokens in the passage sequence $x$ with special tokens {\tt [MASK]}, as opposed to the whole sequences used in standard BERT.
The main difference between O-MLM and MLM is the way of masking. 
We always mask the {\tt [O-MASK]} tokens to predict ``yes'' or ``no'', as shown in Figure~\ref{fig:overview_unimc} (b). 
{Therefore, the loss $\mathcal{L}_{\mathrm{O-MLM}}$ and $\mathcal{L}_{\mathrm{MLM}}$ share the same style.}

Once the prediction probabilities of ``yes'' or ``no'' is obtained, we next introduce the OP to teach the model for learning MC tasks, which is shown in Figure~\ref{fig:overview_unimc} (b). 
To learn the mutually exclusive characteristics between options, OP takes the logits $T_{\tt [O-MASK]}^{yes} \in \{ T_{{\tt [O-MASK]}^1}^{yes}, \dots, T_{{\tt [O-MASK]}^{N_O}}^{yes} \}$ in ``yes'' for each option sequence to generate label distributions. 
OP aims to compute a cross-entropy loss with ground truth label distribution $Y$:
\begin{equation}
\small
\mathcal{L}_{\mathrm{OP}}=-\sum_{i=1}^{N_O} Y_i \log \operatorname{Softmax}\left( T_{\tt [O-MASK]}^{yes}\right)
\end{equation}

\begin{figure}[t]
    \centering
    \includegraphics[width=0.5\textwidth]{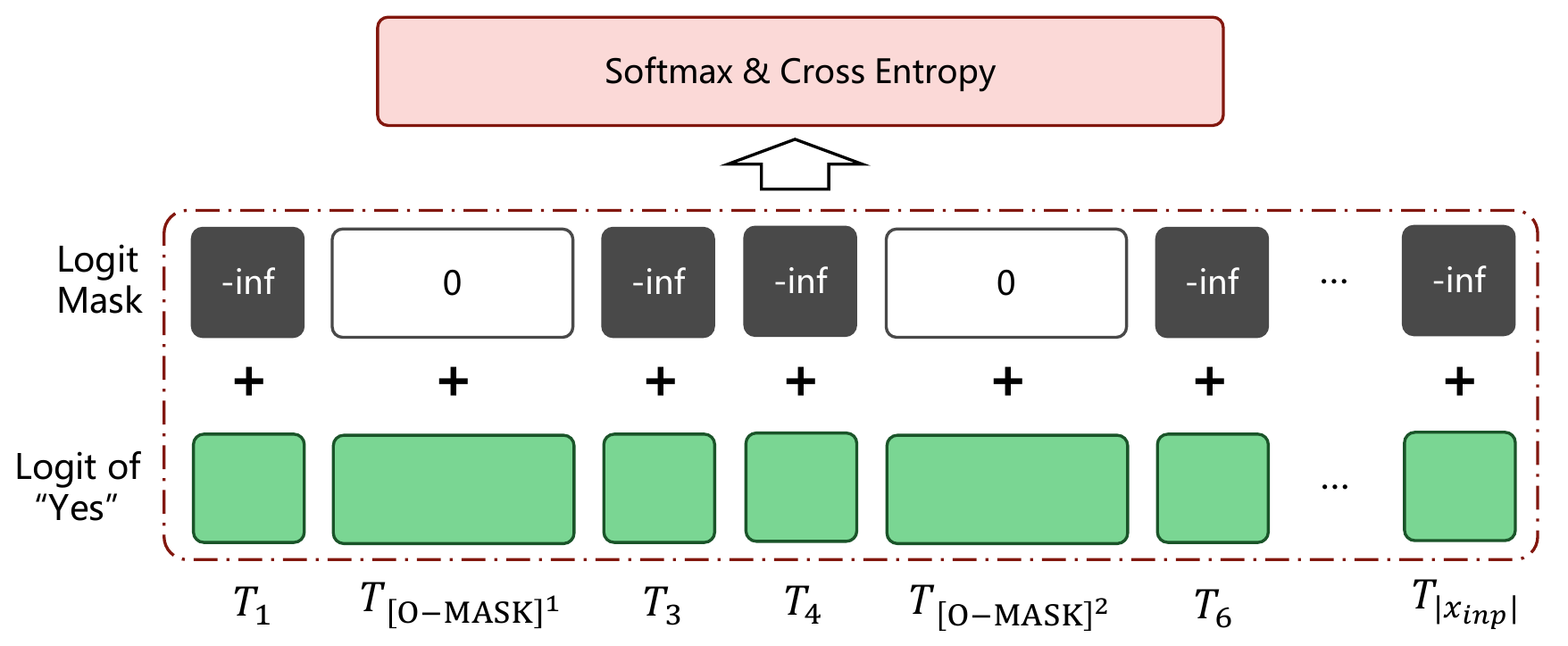}
    \caption{Applying logit masking method in OP. $-inf$ means negative infinity.}
    \label{fig:logit_mask}
\end{figure}

Recent studies show that including mixed tasks in a batch will improve the generalization ability of neural networks~\cite{DBLP:conf/emnlp/AghajanyanGSCZG21/muppet}. 
When facing mixed tasks, we mask the output logits except for {\tt [O-MASK]} during the Softmax operation to compute the OP loss in a mini-batch, as shown in Figure~\ref{fig:logit_mask}.
The logit masking approach allows our UniMC to handle MC tasks with different number of options in a single batch.

In summary, the overall training objective in MC training is given by:
\begin{equation}
\mathcal{L}_{\mathrm{full}}=\mathcal{L}_{\mathrm{MLM}} + \mathcal{L}_{\mathrm{O-MLM}} + \mathcal{L}_{\mathrm{OP}}
\end{equation}

\subsubsection{Zero-shot phase}
\label{sec:zero_shot_phase}

After obtaining a unified MC model, we simply utilize O-MLM and OP to predict the answer in unseen zero-shot datasets. 
We know that the ground truth labels are missing, so it is impossible to compute the loss.
Alternatively, we can compute the most confident option with the OP because the model still recover {\tt [O-MASK]} to ``yes'' or ``no'' with O-MLM.

\subsubsection{Discussion}

Interestingly, we realize that the MC training stage and zero-shot stage are consistent in processing objectives. 
% 因为在zero-shot的时候是没办法计算loss的，也就“无法学习”。所以应该是，数据处理/预测的目标是一致的。
Recall that previous models tend to have divergence learning objectives, which may cause potential oscillation. 
Our proposed method is more task-driven and thus has a better chance to deliver high learning quality in task-specific outputs.

\begin{figure}[tp]
    \centering
    \includegraphics[width=0.48\textwidth]{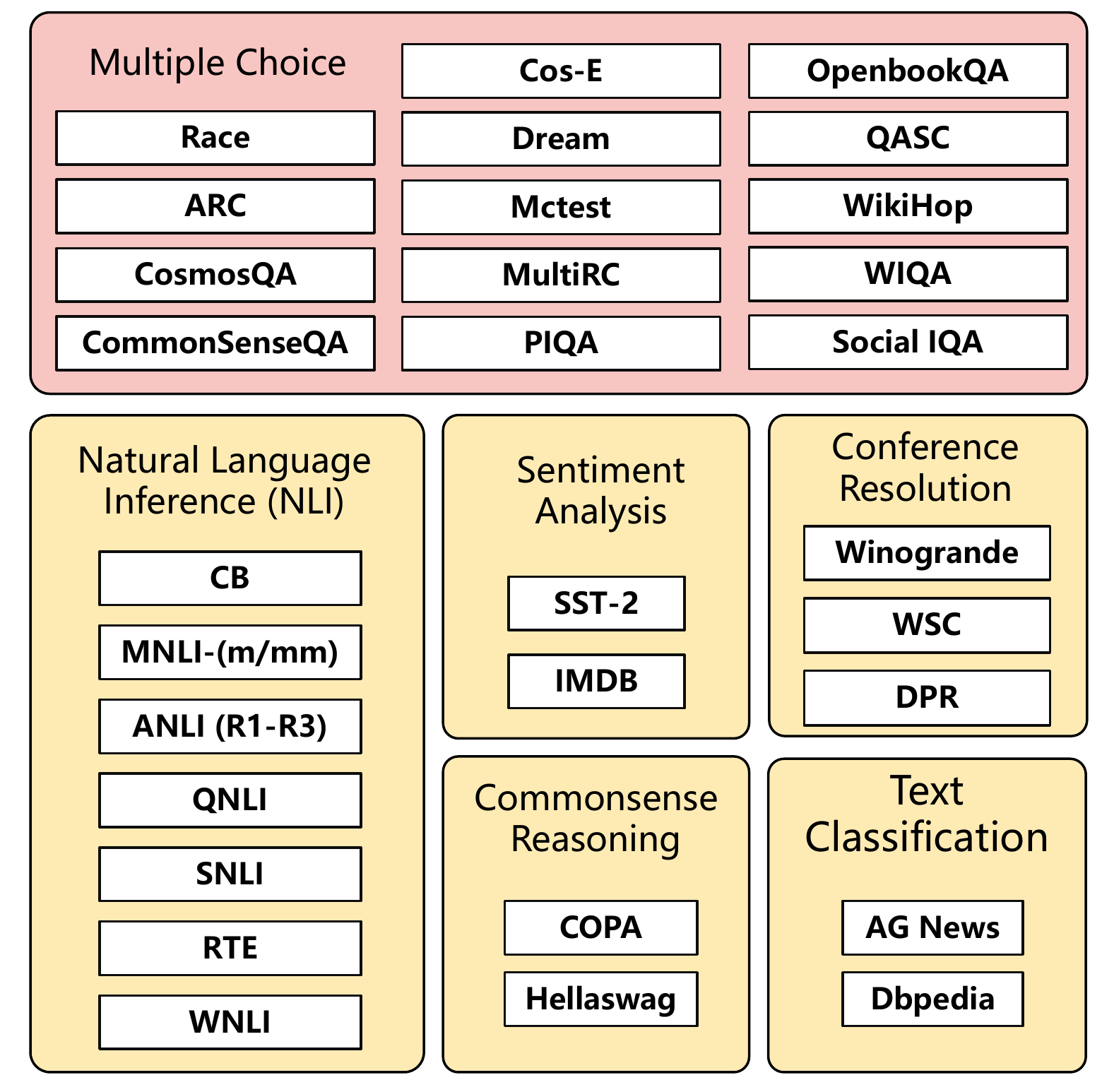}
    \caption{Datasets with various types of tasks. Datasets in MC training phase are in red (above). Datasets in zero-shot phases are in yellow (below).}
    \label{fig:datasets}
\end{figure} 

%% file: sections/04experiment.tex
\subsection{Experimental Setup}

\begin{table*}[!tp]
\begin{center}
\begin{small}
\begin{adjustbox}{max width=\textwidth}
\begin{tabular}{lccccc}
\toprule
\textbf{Model}                                    & \textbf{T0 11B}        & \textbf{GLaM 60B} & \textbf{FLAN 137B} & \textbf{PaLM 540B}   & \textbf{UniMC 235M}    \\ \midrule
Parameters             & $\times$46.8          & $\times$255.3    & $\times$583.0     & $\times$2297.9        & $\times$1.0           \\ \midrule
ANLI R1    & 43.6          & 40.9     & 47.7      & 48.4          & \textbf{52.0} \\
ANLI R2    & 38.7          & 38.2     & 43.9      & 44.2          & \textbf{44.4} \\
ANLI R3    & 41.3          & 40.9     & 47.0      & 45.7          & \textbf{47.8} \\
CB         & 70.1          & 33.9     & 64.1      & 51.8          & \textbf{75.7} \\ \bottomrule
\end{tabular}
\end{adjustbox}
\end{small}
\end{center}
\caption{Zero-shot results in natural language inference task. The best scores are in \textbf{bold}.}
\label{table:nli_results}
\end{table*}

We follow the preparation in T0~\cite{DBLP:journals/corr/abs-2110-08207/T0} to cluster the label-based NLP datasets into $6$ groups. 
In particular, we collect publicly available NLP datasets on HuggingFace\footnotemark[1], and assign each label-based dataset to one of the task groups, as shown in Fig.~\ref{fig:datasets}. 
For each group, we design a corresponding transformation rule to convert it into a unified MC format, where detailed examples are presented in Sec.~\ref{sec:unified_input}. 
Please refer to Appendix~\ref{append:dataset_details} for more details of dataset descriptions and unified MC formats.
Next, we split the whole datasets into two parts for the two phases in our framework, i.e., the part for MC task is for the training, and the other is for the zero-short scenarios. 
It is worthy mentioning that using the MC tasks only in the MC training phase can avoid intensive resource computing. 

\footnotetext[1]{\url{https://huggingface.co/datasets}}

Following the general setting~\cite{DBLP:journals/corr/abs-2112-06905/galm,Wei2021FinetunedLM/FLAN}, we apply accuracy in all datasets. 
For computing the overall average accuracy, we take the average accuracy for each task and then calculate the arithmetic mean for them.

\begin{table}[!tp]
\begin{center}
\begin{small}
\begin{adjustbox}{max width=0.48\textwidth}
\begin{tabular}{lccc}
\toprule
\textbf{Dataset} & \textbf{GPT2} & \textbf{GPT3*} & \textbf{UniMC} \\ \midrule
AG News  & 68.3 & 73.9  & \textbf{81.3}  \\
Dbpedia & 52.5 & 59.7  & \textbf{88.9}  \\ \bottomrule
\end{tabular}
\end{adjustbox}
\end{small}
\end{center}
\caption{Zero-shot results in text classification task. The best results are in \textbf{bold}.}
\label{table:classification_results}
\end{table}

\subsubsection{Baselines}

In the experiments, we compare our method with the state-of-the-art baselines, including: GPT2~\cite{radford2019language/gpt2}, GPT3$^*$~\cite{DBLP:conf/icml/ZhaoWFK021/gpt3c}, T0~\cite{DBLP:journals/corr/abs-2110-08207/T0}, FLAN~\cite{Wei2021FinetunedLM/FLAN}, PaLM~\cite{DBLP:journals/corr/abs-2204-02311/PaLM},  GaLM~\cite{DBLP:journals/corr/abs-2112-06905/galm} and UnifiedQA~\cite{DBLP:conf/emnlp/KhashabiMKSTCH20/unifiedqa}. 
We report the accuracy of each method to measure their performance. 
We only present the average outcomes if the baseline is conducted in multiple runs. 
Besides, we include the random guessing as a naive baseline for the comparison.

\subsubsection{Implementation Details}

In our model, we use the ALBERT-xxlarge-V2~\cite{DBLP:conf/iclr/LanCGGSS20/albert} as backbone models by taking its light-weighted parameters.
For fair comparison, we set the maximum length token as $512$ in all experiments as in~\cite{DBLP:conf/iclr/LanCGGSS20/albert}.
In the training, we run only one epoch by following the setting in FLAN~\cite{Wei2021FinetunedLM/FLAN}.
We set the number of samples for each task up to $20K$, aiming to prevent the model from being dominated by specific tasks. 
Besides, we repeat the experiment $5$ times by using different seeds. 
We run all our experiments on 8 NVIDA A100 GPUs.

\subsection{Main Results}

\subsubsection{Natural Language Inference}

We now present our main results from the Natural Language Inference (NLI) task in Table~\ref{table:nli_results}.
UniMC achieves the best performance in all datasets, demonstrating its capability of NLI. 
In particular, UniMC achieves these competitive results with as few as 235M parameters as opposed to hundred billions of parameters in other baselines.
These results confirm the effectiveness of unifying formats as a multiple choice style. 
Besides, a bi-directional structure in UniMC strengths its ability in capturing information as opposed to the previous one-directional structures.

\subsubsection{Text classification}

Text classification task aims to select a label/class for given texts. 
This is similar to the objective of MC task in nature. 
Therefore, we conduct a zero-shot text classification experiment to verify our model's capability. 
As shown in Table~\ref{table:classification_results}, UniMC outperforms previous SOTA models by a large margin. 
In particular, we know that Dbpedia includes $13$ categories, adding a significant challenge to the classification task.
Fortunately, UniMC has a built-in advantage in dealing with multiple classes due to the similarity between choices and classes, leading up to $48.9\%$ improvement.

\subsubsection{A comprehensive comparison to FLAN}

\begin{table}[tp]
\begin{center}
\begin{small}
\begin{adjustbox}{max width=0.49\textwidth}
\begin{tabular}{lcc|lcc}
\toprule
\textbf{Dataset} & \textbf{FLAN} & \textbf{UniMC} & \textbf{Dataset} & \textbf{FLAN}        & \textbf{UniMC}       \\ \midrule
\multicolumn{3}{l|}{{\ul \textbf{NLI}}}           & \multicolumn{3}{l}{{\ul \textbf{Commonsense}}}                 \\
ANLI R1          & 47.7          & \textbf{52.0}  & COPA             & 90.6                 & \textbf{95.2}        \\
ANLI R2          & 43.9          & \textbf{44.4}  & Hellaswag        & 56.4                 & \textbf{62.5}        \\ \cmidrule{4-6} 
ANLI R3          & 47.0          & \textbf{47.8}  & \multicolumn{3}{l}{{\ul \textbf{Coreference}}}                  \\
CB               & 64.1          & \textbf{75.7}  & Winogrande       & \textbf{67.3}        & 65.8                 \\
RTE              & \textbf{78.3} & 78.1           & WSC              & \textbf{80.8}        & 78.8                 \\
QNLI             & \textbf{59.6} & 54.0           & DPR              & 60.3                 & \textbf{87.5}        \\ \cmidrule{4-6} 
SNLI             & 43.0          & \textbf{60.9}  & \multicolumn{3}{l}{{\ul \textbf{Sentiment}}}                   \\
MNLI-m           & 51.1          & \textbf{52.7}  & SST-2            & \textbf{92.6}        & 91.6                 \\
MNLI-mm          & 51.0          & \textbf{51.4}  & IMDB             & 94.1                 & \textbf{94.8}        \\
WNLI             & 61.0          & \textbf{65.4}  &                  & \multicolumn{1}{l}{} & \multicolumn{1}{l}{} \\ \bottomrule
\end{tabular}
\end{adjustbox}
\end{small}
\end{center}
\caption{A summary on natural language inference, commonsense reasoning, coreference resolution and sentiment analysis task.}
\label{table:unimc_vs_flan}
\end{table}

We know that FLAN is a well-known model in dealing with zero-shot option or label-related tasks. 
One of its particular merits is the zero-shot generalization ability.
To better demonstrate the ability of UniMC, we report a comprehensive comparison between ours and FLAN, as shown in Table~\ref{table:unimc_vs_flan} wjj{and more comparisons are decribed in Appendix~\ref{append:full_results}}.
In the NLI task, UniMC achieves better performance than FLAN in general, which is consistent with the results in Table~\ref{table:nli_results}. 
We also select tasks like the commonsense reasoning, the coreference resolution, and the sentiment analysis to further explore the generalization ability of ours. 
UniMC gets an obvious advantage in COPA, Hellaswag, Winogrande, WSC, DPR when evaluating the common sense and coreference tasks. 
Beyond these two tasks, we find that the construction of datasets plays a critical role to the performance.
In general, these datasets can be grouped into two categories: the understanding and generation styles.
UniMC tends to show better performance on datasets that more close to the understanding style.
In sentiment tasks, the number of classes is limited, making the dataset construction style is less important than that in the tasks of  the common sense and coreference. 
Therefore, both UniMC and FLAN get relative good performance. 

\subsection{Ablation Studies}

In this section, we intend to verify the necessity of key components of our UniMC, including the MC training, the prompt effect, the flow controlling. 
We also show the influence of the model size.

\begin{table}[tp]
\begin{center}
\begin{small}
\begin{adjustbox}{max width=0.48\textwidth}
\begin{tabular}{lccc}
\toprule
\textbf{Task}  & \textbf{Random Guess} & \textbf{UniMC*} & \textbf{UniMC}         \\ \midrule
NLI            & 38.3         & 38.1   & \textbf{58.2}  \\
Commonsense   & 37.5          & 43.2   & \textbf{78.9}  \\
Sentiment      & 50.0         & 40.0   & \textbf{93.2}  \\
Coreference     & 50.0        & 54.8   & \textbf{77.4}  \\
Classification & 16.1         & 15.9   & \textbf{85.1}  \\ \midrule
Average        & 38.4         & 38.4   & \textbf{78.6}  \\ \bottomrule
\end{tabular}
\end{adjustbox}
\end{small}
\end{center}
\caption{MC training improves UniMC zero-shot performance. ``UniMC$^*$'' indicates the UniMC without the MC training stage.}
\label{table:abla_mc_training}
\end{table}

\subsubsection{How important is MC training?}

Recall that our proposed UniMC takes advantage of O-MLM and OP to evaluate zero-shot tasks without MC training. 
To better understand our design, we develop a variant of our model that omits the  MC training, named as UniMC$^*$. 
In Table~\ref{table:abla_mc_training}, we present the results of UniMC$^*$, where its performance is close to ``Random Guess''.
This striking outcome verifies the necessity of MC training.

\begin{table}[!tp]
\begin{center}
\begin{small}
\begin{adjustbox}{max width=0.48\textwidth}
\begin{tabular}{lcc}
\toprule
\textbf{Dataset}                       & \textbf{with Question}        & \textbf{w/o Question}         \\ \midrule
{\ul \textbf{NLI}}            & \multicolumn{1}{l}{} & \multicolumn{1}{l}{} \\
ANLI R1                       & 47.5                 & \textbf{52.0}          \\
ANLI R2                       & 43.2                 & \textbf{44.4}        \\
ANLI R3                       & 46.4                 & \textbf{47.8}        \\
QNLI                          & 52.2                 & \textbf{54.0}          \\
RTE                           & 74.3                 & \textbf{78.1}        \\
WNLI                          & 59.4                 & \textbf{65.4}        \\
MNLI-m                        & \textbf{52.7}        & 48.8                 \\
MNLI-mm                       & \textbf{51.4}        & 47.5                 \\
CB                            & \textbf{75.7}        & 70.7                 \\
SNLI                          & \textbf{60.9}        & 53.7                 \\ \midrule
{\ul \textbf{Sentiment}}      & \multicolumn{1}{l}{} & \multicolumn{1}{l}{} \\
SST-2                         & \textbf{91.6}        & 90.2                 \\
IMDB                          & \textbf{94.8}        & 93.6                 \\ \midrule
{\ul \textbf{Classification}} & \multicolumn{1}{l}{} & \multicolumn{1}{l}{} \\
AG News                        & 81.2                 & \textbf{81.3}        \\
Dbpedia                       & 60.1                 & \textbf{88.9}        \\ \bottomrule
\end{tabular}
\end{adjustbox}
\end{small}
\end{center}
\caption{We report results of UniMC with and without question prompts. We present 3 tasks (NLI, Sentiment, Classification) because question prompts are not designed in other tasks.}
\label{table:abla_question_prompt}
\end{table}

\subsubsection{How does the prompt affect the performance?}
\label{sec:ablation_prompt}

Our framework intends to reduce the effort of designing prompts, we now analyze what the effect of particular prompts, including the question prompts and the option prompts.
We present the results in the Table~\ref{table:abla_question_prompt}.

\begin{table}[!htb]
\begin{center}
\begin{small}
\begin{adjustbox}{max width=0.48\textwidth}
\begin{tabular}{lccccc}
\toprule
\textbf{Dataset} & \begin{tabular}[c]{@{}c@{}}\textbf{Good} /\\ \textbf{Bad}\end{tabular} & \begin{tabular}[c]{@{}c@{}}\textbf{Great} /\\ \textbf{Terrible}\end{tabular} & \begin{tabular}[c]{@{}c@{}}\textbf{Positive} /\\ \textbf{Negative}\end{tabular} & \textbf{Average}       & \textbf{Std}        \\ \midrule
\multicolumn{6}{l}{Model: UnifiedQA-T5-3B}                                                                                                                                                                        \\
SST-2   & 71.0                                               & 83.4                                                     & 91.2                                                        & 81.8          & 8.3       \\
IMDB    & 85.4                                               & 90.3                                                     & 90.6                                                        & 88.8          & 2.4       \\ \midrule
\multicolumn{6}{l}{Model: UniMC-235M}                                                                                                                                                                         \\
SST-2   & 90.9                                               & 91.6                                                     & 91.1                                                        & \textbf{91.2} & {\ul 0.3} \\
IMDB    & 94.3                                               & 94.8                                                     & 93.7                                                        & \textbf{94.2} & {\ul 0.4} \\ \bottomrule
\end{tabular}
\end{adjustbox}
\end{small}
\end{center}
\caption{Zero-shot results in sentiment analysis task. ``Std'' indicates Standard Deviation. The best average results are in \textbf{bold}. The more stable performance is {\ul underlined}.}
\label{table:abla_option_prompt}
\end{table}

\begin{table}[!tp]
\begin{center}
\begin{small}
\begin{adjustbox}{max width=0.48\textwidth}
\begin{tabular}{lcc}
\toprule
\textbf{Method}           & \textbf{Average}       & \textbf{Improve} \\ \midrule
Random Guessing     & 38.4          & $+$0.0  \\
Only UIE         & 39.1          & $+$0.7  \\
Only AMM         & 78.0          & $+$39.6 \\
UIE + AMM & \textbf{78.6}        & $+$40.2 \\ \bottomrule
\end{tabular}
\end{adjustbox}
\end{small}
\end{center}
\caption{Zero-shot performance with different strategies to control the flow between options. ``UIE'' indicates Updating Id Embeddings, including segment id and position id. ``AMM'' means Attention Mask Matrix. ``Improve'' shows the accuracy improvement from Random Guessing.}
\label{table:abla_uie_oma}
\end{table}

For the question prompts, we conduct experiments on four challenge tasks by showing performance of using prompts or not. 
Although the performance for all tasks shows different directions, we hypothesize that this divergence is caused by the way of data construction. 
These datasets are mainly designed for two purposes, which are the language modeling task and the relationship choice task~\cite{DBLP:conf/nips/BrownMRSKDNSSAA20/gpt3}. 
The desire for question prompts increases when the data is more close to the language modeling task; vice versa. 
Furthermore, we classify these datasets into two categories, spoken-based and written-based, according to the definition in~\cite{Alsaawi2019SpokenAW/spoken_written}.
MNLI-m/mm, CB, SNLI, SST-2 and IMDB belong to the spoken-based corpus, while the rest datasets belong to written-based corpus.
Considering that PMLM is usually pre-trained on written-based corpus, e.g., the pre-training datasets of BERT are Wikipeida and BookCorpus~\cite{DBLP:conf/naacl/DevlinCLT19/bert}, ours may have no need of questions for written-based data.
This, again, confirms that data construction affects the requirements of question prompts.

For the option prompts, we present the experimental results in Table~\ref{table:abla_option_prompt}.
We would like to emphasize that option prompts are necessary for our UniMC, therefore, we cannot remove this component as in the above experiment.
Instead, we design different option prompts to demonstrate their effects.
We observe that different prompts show limited performance variations, indicating the robustness of our UniMC to option prompts. 
Since FLAN and PaLM are not open-sourced, we choose one of the most powerful models, e.g., UnifiedQA-T5, as the baseline to ensure the fairness in comparison.
In the experiment, we find that UnifiedQA-T5 is sensitive to option prompts, which have up to $8.3$ standard variation (Std).

\subsubsection{How does the flow controlling affect the performance?}

We design the prompt to frame the input sequences to make all datasets fit into UniMC directly.
However, some recent methods need extra processes, such as adopting an option with a context (question and passage) into a sequence and aggregate multiple different sequences to get an answer~\cite{DBLP:journals/corr/abs-2109-03564/NSP-BERT}. 
To fix this gap, we design two strategies to control the flow of the information as in Section~\ref{sec:backbone_network}.
We summarize the performance of these two in Table~\ref{table:abla_uie_oma}.
We observe that AMM adds the greatest improvement to results, which is much better than UIE.
On the one hand, UniMC can learn the position relationship between options. 
On the other hand, UniMC can distinguish between options and context. 
However, UIE is unable to prevent the inter-influence in options. 
Thanks to self-attention mechanism, AMM makes the options unreachable to each other, eliminating the intra-information of options.

\subsubsection{How does the model size affect the performance?}

\begin{figure}[tp]
    \centering
    \includegraphics[width=0.48\textwidth]{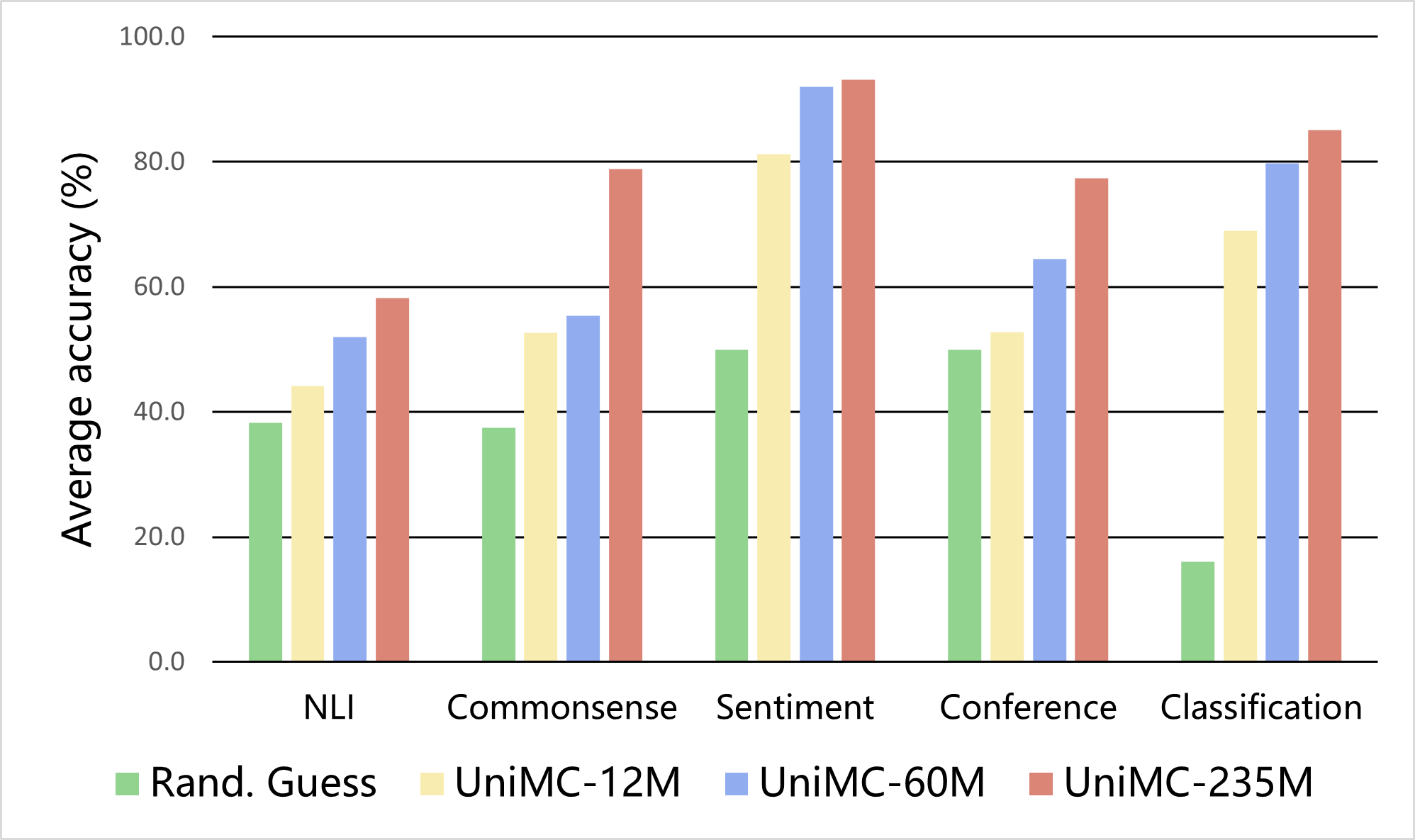}
    \caption{Zero-shot performances on several tasks with model variants.}
    \label{fig:abla_model_size}
\end{figure} 

A common intuition from this domain is that a large model size will result a better performance~\cite{Wei2021FinetunedLM/FLAN,DBLP:journals/corr/abs-2204-02311/PaLM}, particular large-scale PLMs.
Naturally, we believe that our backbone PMLM follows this rule as well. 
To validate this, we implement an experiment by varying the model size, as shown in Figure~\ref{fig:abla_model_size}. 
All $4$ different tasks show the same trend, demonstrating the correctness of the mentioned intuition.

%% file: sections/05conclusion.tex
In this paper, we introduce a new zero-shot paradigm called MC tuning. 
This adds flexibility and generalization ability to zero-shot learners.
We propose O-MLM and OP in both MC training and zero-shot phase, aiming to capture information from both directions.
Our UniMC achieves better performances over SOTA models that a few hundred times larger than our model. 
Our experiments demonstrate the effectiveness and generalization ability of UniMC on zero-shot tasks. 
In future work, we will extend UniMC to few-shot scenarios.

%% file: sections/06limitation.tex
\label{sec:limitation}

In this paper, our main contribution is a simple and effective framework for zero-shot tasks while maintaining a light weight. 
We aim to introduce additional artificial information and reduce manual processing to the minimum. 
We explored how to employ question prompts in Sec.~\ref{sec:ablation_prompt}, however, it is non-trivial to decide whether a prompt is required for complex datasets.
In addition, we only compare with limited baselines when understanding the influence from the backbone in UniMC. 
In experiments, we implement only a few comparative experiments between ALBERT and RoBERTa~\cite{DBLP:journals/corr/abs-1907-11692/roberta} due to the limit of computational resources, as shown in Appendix~\ref{append:different_backbone}. 
In the future, we will dig deeper into the principles regarding inputs and backbone, etc.

%% file: sections/07ethic.tex
Natural language processing is an important technology in our society. 
It is necessary to discuss its ethical influence~\cite{DBLP:conf/ethnlp/LeidnerP17/ethical_desgin}. 
In this work, we develop a novel zero-shot NLP approach to enhance the generalization ability of NLP.
As discussed in~\cite{schramowski2022large/ethic_large_model, DBLP:journals/corr/abs-1912-05238/ethic_bert,DBLP:conf/acl/BlodgettBDW20/ethic_language}, language models might contain human-like biases, which might embed in both the parameters of the models and outputs. 
Furthermore, we note the potential abuse of zero-shot models because these are often being integrated into applications without justification.
We encourage open debating on its utilization, such as the task selection and the deployment, hoping to reduce the chance of any misconduct.

%% file: sections/ack.tex
This research was supported by the open-source project, Fengshenbang~\cite{fengshenbang}. 
We would like to thank members of The Real Sakai Laboratory\footnotemark[2] and of the GTS team in IDEA\footnotemark[3], for giving us suggestions.
Junjie Wang is especially grateful to our friend Yuxiang Zhang for his support, advice, and encouragement.

\footnotetext[2]{\url{http://sakailab.com/english/}}

\footnotetext[3]{\url{https://www.idea.edu.cn/}}

%% file: sections/appendix.tex
\appendix

\section*{Appendix}
\label{sec:appendix}

\section{Dataset Details}
\label{append:dataset_details}

Based on their usage stages, We summarize all datasets in two parts: MC training datasets and evaluation datasets.

\begin{table}[!htbp]
\begin{center}
\begin{small}
\begin{adjustbox}{max width=0.48\textwidth}
\begin{tabular}{lcc}
\toprule
\textbf{Datasets}      & \textbf{\# of option} & \textbf{\# of examples} \\ \midrule
ARC           & $4$             & 3.37k          \\
CommonsenseQA & $5$             & 9.7k           \\
Cos-E         & $5$             & 10.9k          \\
CosmosQA      & $4$             & 25.2k          \\
Dream         & $4$             & 10k            \\
Mctest        & $4$             & 2.4k           \\
MultiRC       & multiple       & 12k            \\
OpenbookQA    & $4$             & 9.9k           \\
PIQA          & $2$             & 16.1k          \\
QASC          & $8$             & 8.1k           \\
Race          & $4$             & 87.8k          \\
Socail IQa      & $3$             & 33.4k        \\
WikiHop       & multiple       & 43.7k          \\
WIQA          & 3             & 36.7k          \\ \bottomrule
\end{tabular}
\end{adjustbox}
\end{small}
\end{center}
\caption{Dataset statistics for Multiple Choice task}
\label{table:append_mc_datasets}
\end{table}

\begin{table*}[!htbp]
\begin{center}
\begin{small}
\begin{adjustbox}{max width=\textwidth}
\begin{tabular}{llccc}
    \toprule         
    \textbf{Task} &  \textbf{Dataset} & \textbf{Passage}  & \textbf{Question} & \textbf{Options} \\
    \midrule
       \multirow{11}{*}{NLI}           & ANLI R1   & \multirow{7}{*}{$x_1$} & \multirow{7}{*}{Base on the paragraph.}               & \multirow{7}{*}{\begin{tabular}[c]{@{}l@{}} {[1]} We can infer that $x_2$; \\ {[2]} We can not infer that $x_2$; \\ {[3]} It is difficult for us to infer $x_2$.\end{tabular}} \\
                                & ANLI R2   &                     &                                                       &                                                                                                                                                            \\
                                & ANLI R3   &                     &                                                       &                                                                                                                                                            \\
                                & CB         &                     &                                                       &                                                                                                                                                            \\
                                & SNLI       &                     &                                                       &                                                                                                                                                            \\
                                & MNLI-m     &                     &                                                       &                                                                                                                                                            \\
                                & MNLI-mm    &                     &                                                       &                                                                                                                                                            \\
                                
                                \cmidrule{2-5}
                                
                                & QNLI       & \multirow{3}{*}{$x_1$} & \multirow{3}{*}{Base on the paragraph.}               & \multirow{3}{*}{\begin{tabular}[c]{@{}l@{}} {[1]} We can infer that $x_2$; \\ {[2]} We can not infer that $x_2$.\end{tabular}}                                          \\
                                & RTE        &                     &                                                       &                                                                                                                                                            \\
                                & WNLI       &                     &                                                       &                                                                                                                                                            \\
                                \midrule
\multirow{2}{*}{Sentiment}      & SST-2      & \multirow{2}{*}{\textit{x} }  & \multirow{2}{*}{What is sentiment of reviews?}        & \multirow{2}{*}{\begin{tabular}[c]{@{}l@{}} {[1]} It’s great; \\ {[2]} It’s terrible.\end{tabular}}                                                               \\
                                & IMDB       &                     &                                                       &                                                                                                                                                            \\
                                \midrule
\multirow{5}{*}{Classification} & AG News   & \textit{x}                    & What is the topic of the news?                        & \begin{tabular}[c]{@{}l@{}} {[1]} World news;\\ {[2]} Sports news;\\ {[3]} Business news;\\ {[4]} Technology news.\end{tabular}                                                                 \\
                                \cline{2-5}
                                & Dbpedia    & \textit{x}                   & What is the topic of the articles?                    & \begin{tabular}[c]{@{}l@{}} {[1]} Company;\\ {[2]} Educational Institution;\\ ...\\ {[13]} Written Work.\end{tabular}                                                         
 \\ \bottomrule
\end{tabular}
\end{adjustbox}
\end{small}
\end{center}
\caption{Prompt designs for all datasets.}
\label{table:append_prompt_desgin}
\end{table*}

\subsection{MC training datasets}

Multiple Choice (MC) task aims to select a right answer from multiple candidate options according to the related questions and passages. 
As shown in Table~\ref{table:append_mc_datasets}, we use the following datasets in MC training phase:

\begin{enumerate}
    \item ARC~\cite{DBLP:journals/corr/abs-1803-05457/arc}
    \item CommonsenseQA~\cite{DBLP:conf/naacl/TalmorHLB19/CommonsenseQA}
    \item Cos-E~\cite{DBLP:conf/acl/RajaniMXS19/cos-e}
    \item CosmosQA~\cite{DBLP:conf/emnlp/HuangBBC19/cosmosqa}
    \item Dream~\cite{DBLP:journals/tacl/SunYCYCC19/dream}
    \item Mctest~\cite{DBLP:conf/emnlp/RichardsonBR13/mctest}
    \item MultiRC~\cite{khashabi-etal-2018-looking/multirc}
    \item OpenbookQA~\cite{DBLP:conf/emnlp/MihaylovCKS18/openbookqa}
    \item PIQA~\cite{DBLP:conf/aaai/BiskZLGC20/piqa}
    \item QASC~\cite{DBLP:conf/aaai/KhotCGJS20/qasc}
    \item Race~\cite{DBLP:conf/emnlp/LaiXLYH17/race}
    \item Socail IQA~\cite{sap-etal-2019-social/social_iqa}
    \item WikiHop~\cite{DBLP:journals/tacl/WelblSR18/wikihop}
    \item WIQA~\cite{DBLP:conf/emnlp/TandonDSCB19/wiqa}
\end{enumerate}

\subsection{Evaluation datasets}

To evaluate zero-shot capability of models, we collect several NLP datasets and group them by tasks. 
The datasets with tasks are following: 

\textbf{Natural language inference} (NLI) is to ascertain whether a ``hypothesis'' with a ``premise'' is true (entailment), false (contradiction), or indeterminate (neutral).

\begin{enumerate}
    \item ANLI (R1-R3)~\cite{nie2019adversarial/anli}
    \item CB~\cite{MarieCatherinedeMarneffe2019TheCI/cb}
    \item SNLI~\cite{DBLP:conf/emnlp/BowmanAPM15/snli}
    \item MNLI-m/mm~\cite{williams-etal-2018-broad/mnli}
    \item QNLI~\cite{rajpurkar-etal-2018-know/qnli}
    \item RTE~\cite{DBLP:conf/mlcw/DaganGM05/rte1,giampiccolo2007third/rte2,DBLP:conf/acl/GiampiccoloMDD07/rte3,DBLP:conf/tac/BentivogliMDDG09/rte4}
    \item WNLI~\cite{10.5555/3031843.3031909/wnli}
\end{enumerate}

\textbf{Commonsense reasoning} (Commonsense) requires the model to draw conclusions based on "common sense" or general information.

\begin{enumerate}
    \item COPA~\cite{DBLP:conf/aaaiss/RoemmeleBG11/copa}
    \item Hellaswag~\cite{DBLP:conf/acl/ZellersHBFC19/hellaswag}
\end{enumerate}

\textbf{Sentiment analysis} (Sentiment) is to classify the polarity of a given text.

\begin{enumerate}
    \item SST-2~\cite{DBLP:conf/emnlp/SocherPWCMNP13/sst-2}
    \item IMDB~\cite{maas-EtAl:2011:ACL-HLT2011/imdb}
\end{enumerate}

\textbf{Coreference resolution} (Coreference) is the process of grouping textual mentions that refer to the same underlying real-world objects.

\begin{enumerate}
    \item Winogrande~\cite{DBLP:conf/aaai/SakaguchiBBC20/winogrande}
    \item WSC~\cite{levesque2012winograd/wsc}
    \item DPR~\cite{DBLP:conf/emnlp/RahmanN12/dpr}
\end{enumerate}

\textbf{Text classification} (Classification) is the task of assigning a label to a given text. 

\begin{enumerate}
    \item AG News~\cite{DBLP:conf/nips/ZhangZL15/agnews}
    \item Dbpedia~\cite{DBLP:journals/semweb/LehmannIJJKMHMK15/dbpedia}
\end{enumerate}

\begin{table}[tp]
\begin{center}
\begin{small}
\begin{adjustbox}{max width=0.48 \textwidth}
\begin{tabular}{lcccc}
\toprule
\textbf{Model}  & \textbf{Layers}   & \textbf{Hidden} & \textbf{Heads}  & \textbf{Parameters} \\
\midrule
UniMC-12M  & 12 & 768  & 12 & 12M        \\
UniMC-60M  & 24 & 2048 & 16 & 60M        \\
UniMC-235M & 12 & 4096 & 64 & 235M       \\
\bottomrule
\end{tabular}
\end{adjustbox}
\end{small}
\end{center}
\caption{The configurations if the UniMC variants.}
\label{table:append_variants}
\end{table}

\begin{table}[tp]
\begin{center}
\begin{small}
\begin{adjustbox}{max width=0.4\textwidth}
\begin{tabular}{lcc}
\toprule
                & \textbf{RoBERTa} & \textbf{ALBERT}  \\ \midrule
Parameters      & 355M             & 235M             \\ \midrule
NLI (Acc)       & 53.0             & \textbf{58.2}    \\
Sentiment (Acc) & 92.8             & \textbf{93.2}    \\ \bottomrule
\end{tabular}
\end{adjustbox}
\end{small}
\end{center}
\caption{Ablation experiments with different backbones. ``RoBERTa'' indicates RoBERTa-large and ``ALBERT'' presents ALBERT-xxlarge-v2.}
\label{table:abla_diff_backbones}
\end{table}

\subsection{Unified input}
\label{append:unified_input_text}

Inspired by template examples in FLAN~\cite{Wei2021FinetunedLM/FLAN}, we design a simple rule to transform the original text to a unified MC format as shown in Table~\ref{table:append_prompt_desgin}. 
In addition, we present two examples:

An example of Social IQA (multiple choice):

\begin{enumerate}
    \item Raw text: \{$x_1$: ``Jesse placed the music sheet in his hands and began to sing a song.'', ``question'': ``What will Jesse want to do next?'', ``option'': [``paint a picture'', ``make a speech'', ``start the song''], ``answer'': ``start the song''\}
    \item Transformed text: ``no paint a picture. no make a speech. yes start the song. What will Jesse want to do next? Jesse placed the music sheet in his hands and began to sing a song.''
    \item Input tokens: {\tt [O-MASK]} paint a picture. {\tt [O-MASK]} make a speech. {\tt [O-MASK]} start the song. What will Jesse want to do next? Jesse placed the music sheet in his hands and began to sing a song.
\end{enumerate}

An example of SNLI (natural language inference):

\begin{enumerate}
    \item Raw text: \{$x_1$: ``A man reads the paper in a bar with green lighting.'',  $x_2$: ``The man is inside.'', ``option'': [``we can infer that'', ``we can not infer that The man is inside.'', ``it is difficult for us to infer that The man is inside.''], ``answer'': ``we can infer that The man is inside.''\}
    \item Transformed text: ``yes we can infer that The man is inside. no we can not infer that The man is inside. no it is difficult for us to infer that The man is inside. Base on the paragraph. A man reads the paper in a bar with green lighting.''
    \item Input tokens: ``{\tt [O-MASK]} we can infer that The man is inside. {\tt [O-MASK]} we can not infer that The man is inside. {\tt [O-MASK]} it is difficult for us to infer that The man is inside. Base on the paragraph. A man reads the paper in a bar with green lighting.''
\end{enumerate}

\begin{table*}[]
\begin{center}
\begin{small}
\begin{adjustbox}{max width=\textwidth}
\begin{tabular}{lcccccccc}
\hline
Dataset                     & GPT3 175B & T0 11B        & GLaM 60B/MoE & FLAN 137B     & PaLM 8B & PaLM 60B & PaLM 540B     & UniMC 235M    \\ \hline
{\ul \textbf{NLI}}          &           &               &              &               &         &          &               &               \\
ANLI R1                    & 34.6      & 43.6          & 40.9         & 47.7          & 34.9    & 36.4     & 48.4          & \textbf{52.0} \\
ANLI R2                    & 35.4      & 38.7          & 38.2         & 43.9          & 35.8    & 37.2     & 44.2          & \textbf{44.4} \\
ANLI R3                    & 34.5      & 41.3          & 40.9         & 47.0          & 34.5    & 36.7     & 45.7          & \textbf{47.8} \\
CB                          & 46.4      & 70.1          & 33.9         & 64.1          & 41.1    & 57.1     & 51.8          & \textbf{75.7} \\
RTE                         & 63.5      & \textbf{80.8} & 68.8         & 78.3          & 54.2    & 67.9     & 72.9          & 78.1          \\
QNLI                        & -         & -             & -            & \textbf{59.6} & -       & -        & -             & 54.0          \\
SNLI                        & -         & -             & -            & 43.0          & -       & -        & -             & \textbf{60.9} \\
MNLI-m                      & -         & -             & -            & 51.1          & -       & -        & -             & \textbf{52.7} \\
MNLI-mm                     & -         & -             & -            & 51.0          & -       & -        & -             & \textbf{51.4} \\
WNLI                        & -         & -             & -            & 61.0          & -       & -        & -             & \textbf{65.4} \\ \hline
{\ul \textbf{Commonsense}} &           &               &              & \textbf{}     &         &          &               &               \\
COPA                        & 91.0      & 90.0          & 90.0         & 90.6          & 86.0    & 93.0     & 93.0          & \textbf{95.2} \\
Hellaswag                   & 78.9      & 33.6          & 77.1         & 56.4          & 68.7    & 79.7     & \textbf{83.4} & 62.5          \\ \hline
{\ul \textbf{Sentiment}}    &           &               &              &               &         &          & \textbf{}     &               \\
SST-2                       & 71.6      & -             & -            & \textbf{92.6} & -       & -        & -             & 91.6          \\
IMDB                        & -         & -             & -            & 94.1          & -       & -        & -             & \textbf{94.8} \\  \hline
{\ul \textbf{Coreference}}   &           &               &              &               &         &          &               & \textbf{}     \\
Winogrande                  & 70.2      & 59.9          & 73.4         & 67.3          & 66.3    & 77.0     & \textbf{81.1} & 65.8          \\
WSC                    & 88.3      & 65.1          & 86.8         & 80.8          & 78.9    & 86.3     & \textbf{89.1} & 78.8          \\
DPR                         & -         & -             & -            & 60.3          & -       & -        & -             & \textbf{87.5} \\ \hline
\end{tabular}
\end{adjustbox}
\end{small}
\end{center}
\caption{Zero-shot performances on different tasks: NLI, Commonsense, Sentiment, and Coreference.}
\label{table:full_results}
\end{table*}

\section{Additional Experiments}
\label{append:experiment_details}

\subsection{UniMC variants with different parameters}
\label{append:unimc_variants}

By following the setting of ALBERT~\cite{DBLP:conf/iclr/LanCGGSS20/albert}, UniMC employs various ALBERT models as the backbones as shown in Table~\ref{table:append_variants}.

\subsection{Further ablation study: Different backbone models}
\label{append:different_backbone}

To explore the effect of different backbone models in UniMC, we replace the ALBERT-xxlarge-v2 with RoBERTa-large. 
As seen in Table~\ref{table:abla_diff_backbones}, ALBERT outperforms RoBERTa in the NLI and sentiment analysis task. 
A simple explanation is that ALBERT-xxlarge-v2~\cite{DBLP:conf/iclr/LanCGGSS20/albert} (88.9 point) performs beyond RoBERTa-large~\cite{DBLP:journals/corr/abs-1907-11692/roberta} in their paper. 
In our experiments, tokenization might be another possible reason. 
Since O-MLM aims to predict ``yes'' or ``no'', UniMC needs a stable tokenizer to recover those words. 
Unlike ALBERT, RoBERTa uses a byte-level BPE tokenizer instead of a WordPiece tokenizer. 
Under the settings of the byte-level BPE tokenizer, the word id does not only depend on the word itself, but also is influenced by its position. 
Therefore, RoBERTa faces tough O-MLM and OP tasks in the MC training phase, which presents lower score than ALBERT. 
We chose ALBERT, which has better results, as the default backbone model in all our experiments.

\subsection{Results on all datasets} 
\label{append:full_results}

In Table~\ref{table:full_results}, we can see that UniMC achieves the best performance on $11$ out of $17$ datasets.
PLMs outperform UniMC in the tasks of commonsense reasoning and coreference resolution in Hallawag, Winogrand, and WSC, as these are formulated in the original language modeling pre-training objective, as noted in~\citep{Wei2021FinetunedLM/FLAN}. 
In addition, PLMs benefit from unsupervised language modeling on a large-scale text corpus. 
For example, PaLM with 540B parameters is pre-trained on data with $780$ billion tokens. 